\pdfoutput=1
\documentclass[letterpaper, 10 pt, conference]{ieeeconf}

\IEEEoverridecommandlockouts
\usepackage{amsmath}
\usepackage{amssymb}
\usepackage{booktabs}
\usepackage{cite}
\usepackage{graphicx}
\graphicspath{{figs/}}
\usepackage{url}
\usepackage{xcolor}
\definecolor{lenslink}{RGB}{20,80,160}
\definecolor{lenscite}{RGB}{20,115,90}
\usepackage{hyperref}
\hypersetup{pdftitle={Localization and Neural Splatting on the Edge: An FPGA Scene Memory for Robots},
  pdfauthor={Haohua Que, Mingkai Liu, Tianle Zhu, Handong Yao},
  colorlinks=true,linkcolor=lenslink,citecolor=lenscite,urlcolor=lenslink,
  bookmarksnumbered=true}
\usepackage{placeins}

\makeatletter
\def\bstctlcite#1{\@bsphack
  \@for\@citeb:=#1\do{%
    \edef\@citeb{\expandafter\@firstofone\@citeb}%
    \if@filesw\immediate\write\@auxout{\string\citation{\@citeb}}\fi}%
  \@esphack}
\makeatother

\newcommand{\tighttablespacing}{%
  \renewcommand{\arraystretch}{0.88}%
  \setlength{\aboverulesep}{0.35ex}%
  \setlength{\belowrulesep}{0.35ex}%
}
\newcommand{\postfloatvspace}{\vspace{-1.05em}}

\title{\LARGE \bf
Localization and Neural Splatting on the Edge: An FPGA Scene Memory for Robots
}

\author{Haohua Que$^{\dagger}$, Mingkai Liu$^{\dagger}$, Tianle Zhu, and Handong Yao$^{*}$%
\thanks{$^{\dagger}$These authors contributed equally. $^{*}$Corresponding author: Handong Yao.}%
\thanks{All authors are with the School of Environmental, Civil, Agricultural and Mechanical Engineering, College of Engineering, University of Georgia, Athens, GA 30602, USA. {\tt\small handong.yao@uga.edu}}%
}

\begin{document}
\bstctlcite{IEEEtran:BSTcontrol}

\maketitle
\thispagestyle{empty}
\pagestyle{empty}

\begin{abstract}
Resource-constrained robots and autonomous vehicles need scene memories that
localize a camera and render nearby views without a desktop GPU. Scene-coordinate
regression (SCR) and 3D Gaussian splatting (3DGS) fit because they avoid an SfM
stage, but ACEsplat-style pipelines are GPU-centered, and moving them to an edge
FPGA is hard because SCR is pose-sensitive and 3DGS rendering is irregular and
capacity-sensitive. We present LENS, an edge scene-memory acceleration system on
the AMD KR260 that brings an ACEsplat-style workload to a low-power FPGA/ARM
platform. Rather than proposing a new scene model, LENS co-designs the two
dominant kernels with the board data path: an INT16 SCR accelerator that
preserves pose accuracy, and a bit-exact fixed-point ordered 3DGS renderer that rewrites
alpha transmittance as a log-sum-exp recurrence and removes provably useless
streamed Gaussians. All LENS latency, power, and quality results are board-measured.
The SCR engine preserves DSAC* localization accuracy at
25.42 ms, the on-board renderer reaches 34.11 dB on Cambridge against ACEsplat's
33.20, novel-view synthesis on RealEstate10K reaches 27.84 dB against 26.11, and
all ten Wayspots scenes reach the ACEsplat rendering target. LENS wins performance
per watt over a Jetson Orin NX on every deployment task, by 1.9x to 5.7x at about a
quarter of the power, and is faster in absolute latency on localization and
Wayspots rendering. On-device training reaches ACEsplat-level quality but not the
GPU's time or energy, marking the remaining boundary for a future training
accelerator.
\end{abstract}

\begin{figure*}[tp]
    \centering
    \includegraphics[width=\textwidth]{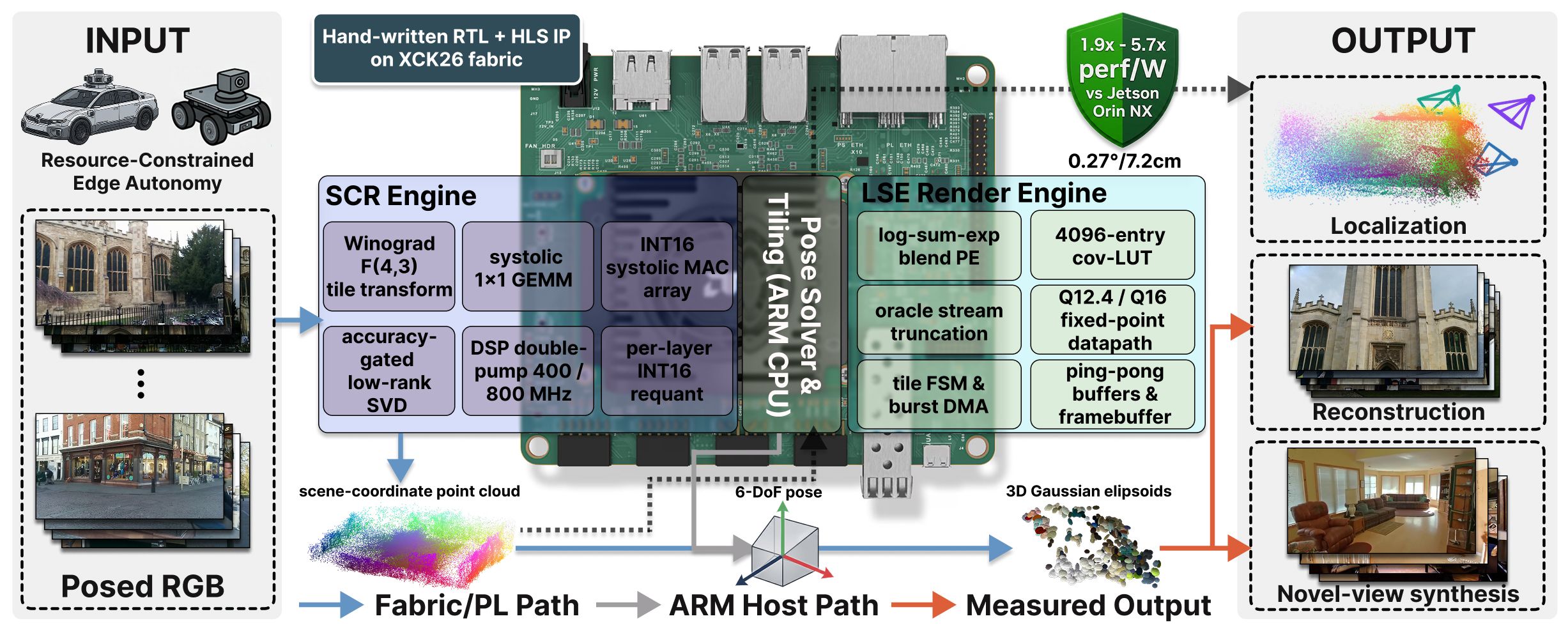}
    \caption{LENS system overview. Posed RGB frames enter the PL SCR engine,
    which produces scene-coordinate point clouds for CPU pose solving and
    Gaussian initialization. The ARM CPU performs pose solving and tiling, while
    the PL LSE render engine rasterizes fixed-point Gaussian streams for
    reconstruction and novel-view synthesis. All outputs are board-measured on
    the KR260.}
    \label{fig:system}
    \vspace{-5pt}
\end{figure*}

\section{Introduction}

Edge autonomy is increasingly limited by what a system can remember under its
own power and thermal budget. A mobile robot revisiting a building, or an
autonomous vehicle returning to a mapped route, needs more than a single pose
estimate. It needs a scene memory that can answer two questions locally: where
is the camera in this place, and what would the scene look like from a nearby
view? Visual localization has been studied through direct pose regression,
prioritized 2-D/3-D search, dense matching, hierarchical retrieval and matching,
and SfM-backed maps
\cite{kendall2015posenet,sattler2017active,brahmbhatt2018mapnet,taira2018inloc,sarlin2019hloc,schonberger2016sfm}.
Today, however,
localization plus rendering is usually built around GPU-backed reconstruction
and neural scene optimization. Such systems are powerful, but they are difficult
to carry into small platforms that must localize, render, and reason while
sharing watts with sensing, planning, and control.

SfM-free neural scene pipelines offer a promising alternative. Scene-coordinate
regression (SCR) predicts dense 3-D coordinates from image pixels
\cite{shotton2013scrf,brachmann2017dsac,brachmann2018dsacpp,brachmann2021dsacstar}.
ACE makes this per-scene relocalization practical within minutes
\cite{brachmann2023ace}. On the rendering side, neural radiance fields
\cite{mildenhall2020nerf,barron2022mipnerf360} and 3D Gaussian splatting (3DGS)
\cite{kerbl20233dgs} have made view synthesis a useful scene-memory interface,
and recent Gaussian SLAM systems use explicit Gaussians for tracking and mapping
\cite{matsuki2024gaussianslam,keetha2024splatam}. Autonomous-driving scene
models have also adopted Gaussians for surround-view dynamic reconstruction and
urban scene understanding
\cite{zhou2024drivinggaussian,yan2024streetgaussians,zhou2024hugs}. ACEsplat
combines SCR and 3DGS into a pipeline for localization, reconstruction, and
novel-view synthesis without an offline SfM stage \cite{acesplat2026}. The
algorithmic story is attractive for autonomy: the same scene representation can
support both relocalization and visual prediction.

The deployment story is harder. General FPGA DNN frameworks show that
quantized neural inference can be efficient on reconfigurable logic
\cite{venieris2016fpgaconvnet,sharma2016dnnweaver,umuroglu2017finn,wang2017pipecnn},
and recent neural-rendering accelerators target NeRF and 3DGS kernels
\cite{wang2023moth,lee2024gscore,wang2025famers}.
LENS targets a different composition of constraints. SCR is a dense CNN, but its
output is geometry, not a class score; small coordinate errors can become large
pose errors, so the convenient INT8 path used by many embedded neural
accelerators is not automatically safe. The 3DGS renderer has the opposite
structure: it is irregular, view-dependent, and dominated by sorted per-pixel
Gaussian contributions. A small tile capacity makes the renderer fast but inaccurate;
a faithful renderer must carry long Gaussian tails through a low-power memory
system. For edge autonomy, success is therefore a system property: pose accuracy,
render quality, latency, power, and board data movement must hold together. The
question is not simply whether ACEsplat can run on an FPGA, but which parts of a
neural scene-memory pipeline can be made accurate, which parts can be made fast,
and which limits remain visible only on the board.

We answer this question with LENS (Localization and Embedded Neural Splatting),
an embedded-FPGA scene-memory system for resource-constrained edge autonomy.
LENS runs on an AMD Kria KR260, a ${\sim}5$ W-class Zynq UltraScale+ platform
\cite{xilinxkr260}, and compares against a Jetson Orin NX edge GPU. It keeps SCR
accuracy-first through INT16 custom logic and moves 3DGS rendering into a bit-exact
fixed-point log-sum-exp datapath. This paper makes a hardware-system claim: an
SfM-free localization and rendering pipeline can serve as an edge scene memory
only when arithmetic, representation, and board-level data movement are
co-designed.
Our contribution is threefold:
\begin{enumerate}
    \item \textbf{A board-measured edge scene-memory system.} We map SCR
    relocalization and 3D Gaussian rendering to a KR260-class FPGA/ARM system
    and report accuracy, latency, and power with explicit board/GPU/emulator
    provenance.
    \item \textbf{An accuracy-gated SCR accelerator.} We show that INT16 custom
    logic, low-rank encoder decomposition, and DSP double-pumping can preserve
    DSAC* pose accuracy while reaching 25.42 ms on board.
    \item \textbf{A bit-exact fixed-point LSE 3DGS render datapath.} We replace the
    loop-carried alpha update with a fixed-point log-transmittance add, add a
    provably bit-exact stream truncation that removes 24 to 60 percent of
    streamed Gaussians, and reach ACEsplat-level reconstruction, novel-view
    synthesis, and Wayspots rendering on board at a performance-per-watt
    advantage over a Jetson Orin NX.
\end{enumerate}

\section{Related Work}

\noindent\textbf{Visual localization and scene coordinates.}
Visual relocalization spans direct pose regression, structure-based 2-D/3-D
matching, visual odometry/SLAM, and SfM-backed maps
\cite{kendall2015posenet,sattler2017active,brahmbhatt2018mapnet,taira2018inloc,sarlin2019hloc,sarlin2020superglue,murartal2015orbslam,murartal2017orbslam2,engel2018dso,teed2021droidslam,schonberger2016sfm,lindenberger2021pixsfm}.
These methods differ in how much geometry is stored explicitly and how much
query-time matching is required. Scene-coordinate methods predict dense
correspondences and solve pose with robust geometry, from regression forests and
DSAC/DSAC++ to DSAC* and ACE
\cite{shotton2013scrf,brachmann2017dsac,brachmann2018dsacpp,brachmann2021dsacstar,brachmann2023ace}.
LENS keeps this relocalization model and studies the hardware mapping when pose
accuracy, not classification accuracy, is the acceptance criterion.

\noindent\textbf{Neural scene representations.}
Neural scene representations have moved from NeRF-style implicit fields to
factorized, baked, voxel, and hash-grid variants
\cite{mildenhall2020nerf,barron2022mipnerf360,chen2022tensorf,yu2021plenoctrees,reiser2021kilonerf,garbin2021fastnerf,hedman2021snerg,fridovichkeil2022plenoxels,muller2022instantngp}
and to explicit 3D Gaussian splatting \cite{kerbl20233dgs}. Recent Gaussian SLAM and
driving-scene systems use Gaussians for tracking, mapping, and urban scene
memory \cite{sucar2021imap,zhu2022niceslam,matsuki2024gaussianslam,keetha2024splatam,zhou2024drivinggaussian,yan2024streetgaussians,zhou2024hugs}.
ACEsplat is closest to our workload because it couples ACE-style SCR with 3DGS
reconstruction and NVS \cite{acesplat2026}, while MACE accelerates coordinate
encoding \cite{mace}. LENS is complementary: it keeps the scene model but moves
its relocalization and rendering bottlenecks onto an embedded FPGA, where the
limits come from arithmetic width, sorted Gaussian streams, and board data
movement rather than from the scene representation itself.

\noindent\textbf{Embedded acceleration for scene memory.}
FPGA DNN accelerators exploit quantization and streaming dataflow
\cite{venieris2016fpgaconvnet,sharma2016dnnweaver,umuroglu2017finn,wang2017pipecnn},
and neural-rendering accelerators target NeRF or 3DGS kernels
\cite{wang2023moth,lee2024gscore,wang2025famers,icarus,instant3d,gsarch,react3d}.
Most accelerate a single render or train kernel, often as ASICs, while a deployed
scene memory must preserve pose accuracy, render quality, latency, and power on
the same board. LENS therefore reports a board-measured localization and
Gaussian-rendering path on a commodity FPGA/ARM platform, exposing both the
regular SCR arithmetic wins and the irregular Gaussian capacity limits.

\section{Workload and System Partition}

LENS targets the ACEsplat-style deployment workload, not a new scene model. The
workload exposes one localization service and two rendering services. For
localization, an ACE-style SCR network predicts a dense 3D scene coordinate for
each image pixel, and a DSAC*-style robust PnP solver estimates the 6-DoF
camera pose from the resulting 2-D/3-D correspondences. For reconstruction and
novel-view synthesis, scene-coordinate predictions initialize Gaussian centers;
the Gaussian colors, opacities, covariances, and rotations define the explicit
scene field rendered from a requested view.

The two kernels pull in opposite directions. SCR is a regular CNN, but its
output is metric geometry, so small coordinate errors can move PnP hypotheses
across inlier thresholds. 3DGS rendering is irregular, depth-ordered,
view-dependent, and capacity-sensitive; dense scenes carry long per-tile Gaussian
tails. LENS therefore has to preserve pose accuracy and ordered Gaussian
compositing within an edge power envelope.

Fig.~\ref{fig:system} shows the KR260 FPGA/ARM partition \cite{xilinxkr260}.
SCR and splatting arithmetic live in the FPGA fabric; orchestration, PnP, tile
binning, depth sorting, and DMA setup remain on the ARM cores. The measured
outputs are the three scene-memory services: localization, reconstruction, and
novel-view synthesis.

\section{Accelerator Design}

\subsection{SCR Accelerator}

The SCR Engine of Fig.~\ref{fig:system} is built in three steps: keep the
arithmetic wide enough to preserve the pose (INT16), make that width fast on the
fabric (a systolic MAC array with DSP double-pumping), then cut the MAC budget
without touching accuracy (accuracy-gated low rank).

\noindent\textbf{INT16 scene-coordinate regression.}
The SCR accelerator runs the full encoder and per-scene head in custom INT16
logic. Each layer quantizes activations and weights with a per-tensor symmetric
scale $s$,
\begin{equation}
\hat{x} = s \cdot \mathrm{clip}\!\left(\mathrm{round}(x/s),\, -2^{15},\, 2^{15}-1\right),
\label{eq:quant}
\end{equation}
and a fixed-point requantization rescales each layer output to the next layer's
input scale. The critical design choice is the 16-bit width: the SCR output is a
3D coordinate, not a class score, and a small coordinate error propagates into a
large pose error through the solver, so the INT8 path used by many embedded
accelerators is not safe here. The 3x3 convolutions run as a Winograd $F(4,3)$
transform,
\begin{equation}
Y = A^{\top}\!\left[\left(G\,g\,G^{\top}\right) \odot \left(B^{\top} d\, B\right)\right] A,
\label{eq:winograd}
\end{equation}
where $g$ is a $3\times3$ tile, $d$ a $6\times6$ input tile, and $A,B,G$ are the
fixed Winograd matrices; this cuts the multiplies per output tile from 36 to 16.
The 1x1 layers map to a GEMM-like systolic array. All arithmetic is bit-exact to
the INT16 reference, so the on-board coordinates match the software coordinates
and preserve DSAC* pose accuracy.

\noindent\textbf{RTL systolic engine and double pumping.}
The HLS implementation exposed a fill/drain and memory-banking wall. We
therefore implemented an output-stationary RTL systolic array (the systolic PE
inset in Fig.~\ref{fig:system}). Each PE accumulates
\begin{equation}
M_{o,p} = \sum_i U_{o,i} V_{i,p},
\end{equation}
with skewed activation and weight streams. A double-pumped variant runs the
DSP48E2 MACs at twice the fabric clock and consumes two input-channel products
per fabric cycle, so the effective throughput of an $N$-PE array is
\begin{equation}
\Theta = 2 \, N \, f_{\mathrm{fabric}} \quad\text{MACs/s},
\label{eq:doublepump}
\end{equation}
at $f_{\mathrm{fabric}}=400$ MHz with the DSPs clocked at $800$ MHz. The
fabric-side feeder, BRAMs, and FSM remain at the slower clock, while a
phase-aligned clock and first-product load control preserve bit-exact
accumulation, so double-pumping doubles throughput without changing the result.

\noindent\textbf{Accuracy-safe low rank.}
To reduce MAC count without retraining the frozen encoder, we decompose the six
dominant 3x3 convolution weight matrices $W \in \mathbb{R}^{C_o \times C_i k^2}$
by a truncated SVD,
\begin{equation}
W \approx \sum_{j=1}^{r} \sigma_j\, u_j v_j^{\top}, \qquad
\rho = \frac{r\,(C_o + C_i k^2)}{C_o\, C_i k^2},
\label{eq:lowrank}
\end{equation}
where $\rho$ is the resulting MAC ratio. A uniform rank ratio of 1.8 gives an
overall $1.63\times$ MAC reduction while preserving localization accuracy under
DSAC* (Table~\ref{tab:scr_speed}). We reject more aggressive compression because
it degrades the harder scenes first, which is the accuracy-gating principle: the
rank is chosen from the DSAC* pose error, not from a fixed compression target.

\subsection{LSE 3DGS Renderer}

The LSE Render Engine of Fig.~\ref{fig:system} is built to carry the ordered
Gaussian stream faithfully yet cheaply: remove the loop-carried timing wall with a
log-sum-exp recurrence, replace every transcendental with a fixed-point table read
so the datapath is exact, and stream the sorted Gaussians through a tile FSM and
burst DMA while dropping the provably useless tail.

\noindent\textbf{From alpha recurrence to log-sum-exp.}
For a pixel $\mathbf{u}$, each projected Gaussian $k$ contributes a weight from
its 2D screen-space footprint,
\begin{equation}
w_k = o_k \, \exp\!\left(-\tfrac{1}{2}\, \mathbf{d}_k^{\top} \Sigma_k^{-1} \mathbf{d}_k\right),
\qquad \mathbf{d}_k = \mathbf{u} - \boldsymbol{\mu}_k,
\label{eq:weight}
\end{equation}
with opacity $o_k$, screen center $\boldsymbol{\mu}_k$, and conic $\Sigma_k^{-1}$.
Standard front-to-back compositing maintains transmittance $T$ and updates color
with $C \leftarrow C + w_k T c_k$ and $T \leftarrow T(1-w_k)$. Directly
implementing this recurrence creates a timing wall because a multiply,
subtraction, and subsequent multiply sit on the loop-carried path. We instead
store $\log T$:
\begin{align}
\log T_{k+1} &= \log T_k + \log(1-w_k), \label{eq:logt}\\
T_k &= \exp(\log T_k), \\
C_{k+1} &= C_k + w_k T_k c_k. \label{eq:color}
\end{align}
The loop-carried path in \eqref{eq:logt} becomes a saturated add, while the
exponential lookup, weight multiplication, and color MACs are feed-forward
pipeline stages. This keeps exact front-to-back ordering and recovers early
stopping when the transmittance falls below a threshold $\tau$, i.e. once
$\log T_k < \log \tau$ the pixel is final.

\noindent\textbf{Fixed-point geometry and tile format.}
The final renderer uses Q12.4 sub-pixel Gaussian centers, Q16 conics, a
4096-entry covariance LUT, Q8 opacity weights, Q12 transmittance, and full-width
color accumulation followed by a single final shift. The Gaussian exponent in
\eqref{eq:weight} is not evaluated in hardware; the quadratic form indexes the
covariance LUT,
\begin{equation}
\begin{aligned}
w_k &= \big(o_k \cdot \mathrm{covLUT}[\,q_k\,]\big) \gg 8, \\
q_k &= \mathrm{clip}\!\left(\lfloor \mathbf{d}_k^{\top}\Sigma_k^{-1}\mathbf{d}_k \cdot \kappa \rfloor,\, 0,\, 4095\right),
\end{aligned}
\label{eq:covlut}
\end{equation}
where $\kappa$ is the fixed covariance scale (300) and $\mathrm{covLUT}$ stores
the exponential decay. This turns the per-pixel weight into two integer reads and
a shift. Gaussian lists are sorted front-to-back on ARM and streamed tile by tile
through the FPGA; each Gaussian is packed as nine INT16 fields and transferred as
five 32-bit words.

\noindent\textbf{On-board render top.}
The render top integrates AXI-Lite control, AXI master DMA through a PS HP port,
per-tile count and offset streams, ping-pong Gaussian buffers, the render array,
and tile-major framebuffer writeback.

\noindent\textbf{Zero-contribution stream compaction.}
The per-pixel blender stops once transmittance saturates, so a Gaussian that
arrives after every pixel in its tile is saturated contributes nothing, yet the
DMA still streams it, and streaming is latency. Let $\mathcal{A}_k$ be the set of
pixels still alive when Gaussian $k$ arrives. We keep $k$ in the tile stream only
if it has a positive integer weight on some live pixel,
\begin{equation}
\mathrm{keep}(k) \;=\; \exists\, \mathbf{u} \in \mathcal{A}_k : w_k(\mathbf{u}) > 0,
\label{eq:oracle}
\end{equation}
with $w_k$ from \eqref{eq:covlut}; the rest are dropped. This is bit-exact because
$w_k=0$ leaves both $\log T$ and $C$ unchanged, and saturated pixels have already
left $\mathcal{A}$. We verified full-framebuffer equality before and after
compaction. The keep-test uses only the already visited depth-sorted stream, so
it is deployable in the ARM tile-preparation pass rather than as an offline
oracle. It removes 24 to 60 percent of streamed Gaussians
(Fig.~\ref{fig:analysis}b), and latency follows the retained-slot count almost
linearly.

\section{Experiments}

\subsection{Experimental Setup}

We evaluate the on-board inference and rendering path for relocalization,
reconstruction rendering, and NVS, and report training only as a boundary
experiment. LENS runs on a KR260 (XCK26, 1248 DSP, four A53 cores, 4 GB DDR4);
the edge baseline is a Jetson Orin NX running the same FP32 SCR and reference
diff-Gaussian renderer. Localization uses DSAC* on 30 frames per Cambridge scene
and representative Wayspots scenes; NVS uses two RealEstate10K~\cite{re10k} input views and
held-out targets. KR260 power comes from the INA260 sensor and Orin power from
tegrastats. Reported latencies are accelerator compute times: the double-pumped
SCR GEMM forward pass and the PL rasterizer at $854\times480$, $16\times16$
tiles, and a 4608-Gaussian cap. ARM PnP, sorting, and tile binning are excluded,
matching the Orin network/render-kernel baseline. The board render is bit-exact
to the integer emulator, so an emulator image and its board framebuffer have the
same metric; Table~\ref{tab:resource} lists post-route resource use.

\begin{table}[tbp]
\caption{Post-route XCK26 utilization (Vivado 2022.2). Both engines meet timing;
the render array is the dominant DSP consumer.}
\label{tab:resource}
\centering
\footnotesize
\setlength{\tabcolsep}{3pt}
\tighttablespacing
\resizebox{\columnwidth}{!}{%
\begin{tabular}{lccccc}
\toprule
Engine & LUT & FF & BRAM & DSP & Clock \\
\midrule
SCR GEMM (2$\times$-pump) & 2126 & 2044 & 0  & 5   & 400/800 MHz \\
LSE render array          & 32010 & 24063 & 24 & 672 & 300 MHz \\
\midrule
XCK26 available           & 117120 & 234240 & 144 & 1248 & \\
\bottomrule
\end{tabular}%
}
\end{table}

Cambridge reconstruction is static-view reconstruction of captured input views,
not held-out NVS. Each per-view field is initialized from SCR coordinates and
optimized only to the first iteration where the board integer render reaches the
scene's ACEsplat PSNR. This avoids over-optimizing into a memorization regime,
and the board, not the emulator, decides the crossing because the emulator
over-reports dense scenes until the hardware tile cap is modeled.

\begin{table*}[tp]
\caption{Deployment summary. Quality is PSNR except localization
(rot\textdegree/trans cm or acc.@10\,cm/5\textdegree); power is LENS/Orin watts.
ACE/Orin NX is matched FP32 on Jetson when available. Training is outside
deployment perf/W claims.}
\label{tab:master}
\centering
\footnotesize
\setlength{\tabcolsep}{3pt}
\tighttablespacing
\resizebox{0.96\textwidth}{!}{%
\begin{tabular}{@{}llccccccc@{}}
\toprule
Task & Dataset & ACE target & LENS & ACE/Orin NX & LENS lat. & Orin NX lat. & Power (W) & perf/W \\
\midrule
Localization    & Cambridge     & 0.42 / 24.9 & 0.44 / 25.2 & 0.42 / 25.8 & \textbf{25.4 ms}  & 33.0 ms  & 3.5 / 14.0  & \textbf{5.2x} \\
Localization    & Wayspots rep. & 63.8\%      & 65.6\%      & 63.3\%      & \textbf{25.4 ms}  & 33.0 ms  & 3.5 / 14.0  & \textbf{5.2x} \\
Reconstruction  & Cambridge     & 33.20       & 34.11       & 34.45       & 31.9 ms  & 16.5 ms  & 4.0 / 14.4  & \textbf{1.9x} \\
Rendering       & Wayspots      & 29.11       & 31.06       & --          & \textbf{10.7 ms}  & 16.5 ms  & 3.9 / 14.4  & \textbf{5.7x} \\
NVS             & RealEstate10K & 26.11       & 27.84       & 28.21       & 31.0 ms  & 16.5 ms  & 4.0 / 14.4  & \textbf{1.9x} \\
\midrule
Training        & Cambridge     & 33.20       & 34.96       & 35.24       & 11.3 h   & 41.5 s   & 4.0 / 12.6  & 0.003x \\
\bottomrule
\end{tabular}%
}
\vspace{-2pt}
\end{table*}

\subsection{Localization and SCR Efficiency}

Localization is the accuracy gate for the whole pipeline: if scene coordinates
move the DSAC* hypotheses across inlier thresholds, a downstream renderer cannot
repair the pose. Table~\ref{tab:scr_loc} shows that the INT16 board coordinates
track FP32 on all Cambridge scenes and on the representative Wayspots scenes. The
mean Cambridge error is 0.44 degrees and 25.2 cm, matching the FP32 reference and
the ACEsplat target.

\begin{table}[tbp]
\caption{Localization accuracy of the INT16 SCR engine plus DSAC* against the FP32
software reference and ACEsplat. Cambridge: median rotation degrees / translation
cm over 30 frames per scene. Wayspots: accuracy at 10\,cm/5\textdegree. The INT16
board result tracks FP32 on every scene.}
\label{tab:scr_loc}
\centering
\setlength{\tabcolsep}{3pt}
\tighttablespacing
\begin{tabular*}{\columnwidth}{@{\extracolsep{\fill}}lccc@{}}
\toprule
Scene & ACEsplat & FP32 & LENS \\
\midrule
\multicolumn{4}{l}{\textit{Cambridge: median rot\textdegree\ / trans cm}}\\
ShopFacade    & 0.30 / 5.0  & 0.27 / 7.1  & 0.27 / 7.2 \\
KingsCollege  & 0.40 / 27.8 & 0.48 / 28.6 & 0.48 / 32.5 \\
OldHospital   & 0.56 / 28.9 & 0.60 / 33.9 & 0.58 / 28.6 \\
StMarysChurch & 0.66 / 20.8 & 0.59 / 18.4 & 0.69 / 19.0 \\
GreatCourt    & 0.19 / 42.2 & 0.18 / 40.8 & 0.16 / 38.6 \\
\cmidrule(lr){1-4}
mean          & 0.42 / 24.9 & 0.42 / 25.8 & 0.44 / 25.2 \\
\midrule
\multicolumn{4}{l}{\textit{Wayspots: acc @10\,cm/5\textdegree}}\\
therock  & 100.0\% & 100.0\% & 100.0\% \\
map      & 56.5\%  & 56.7\%  & 56.7\% \\
tendrils & 34.9\%  & 33.3\%  & 40.0\% \\
\bottomrule
\end{tabular*}
\vspace{-3pt}
\end{table}

\begin{table}[tbp]
\caption{SCR accuracy gates. Top: low-rank ablation on the encoder, with MAC
reduction versus mean Cambridge localization. Bottom: precision gate for the SCR
coordinate field over 103 relocalization frames. The accepted design is the last
compression/precision point that preserves pose accuracy.}
\label{tab:scr_speed}
\centering
\setlength{\tabcolsep}{3pt}
\tighttablespacing
\begin{tabular*}{\columnwidth}{@{\extracolsep{\fill}}lcc@{}}
\toprule
\multicolumn{3}{l}{\textit{Low-rank gate: Cambridge mean localization}}\\
\midrule
Rank ratio & MAC red. & mean (deg / cm) \\
\midrule
1.0 (full)     & 1.00x & 0.42 / 25.3 \\
1.4            & 1.33x & 0.43 / 24.9 \\
1.6            & 1.49x & 0.43 / 24.7 \\
\textbf{1.8}   & \textbf{1.63x} & \textbf{0.43 / 26.8} \\
2.2            & 1.90x & 0.52 / 33.2 \\
2.6            & 2.15x & 0.63 / 38.9 \\
\midrule
\multicolumn{3}{l}{\textit{Precision gate: 103 relocalization frames}}\\
\midrule
Precision & coord cos. & acc. @10\,cm/5\textdegree \\
\midrule
FP32           & 1.000 & 34.0\% \\
\textbf{INT16 (ours)} & \textbf{1.000} & \textbf{35.0\%} \\
INT8           & 0.905 & 4.9\% \\
\bottomrule
\end{tabular*}
\postfloatvspace
\end{table}

The latency improvements in Fig.~\ref{fig:analysis_lines}a are only accepted when the
pose remains within the FP32 envelope. The precision gate in
Table~\ref{tab:scr_speed} explains why: the SCR output is geometry, not a class
label. INT8 coordinate fields have cosine 0.905 versus FP32 and collapse the
10 cm/5 degree accuracy from 34.0 to 4.9 percent. INT16 stays at FP32-level
coordinate cosine and preserves the pose, so the rank and datapath choices are
gated by localization accuracy rather than by compression alone.

\subsection{Cambridge Reconstruction}

With pose accuracy established, Cambridge static-view reconstruction stresses
the rendering path. This experiment uses dense per-view Gaussian fields and
therefore the fixed-point precision, tile capacity, and Gaussian stream bandwidth
of the renderer.

\begin{table}[tbp]
\caption{Cambridge static-view reconstruction. Top: KR260 LSE renderer, three-view
mean at the crossing point; perf/W is relative to Orin NX. Bottom: reference-view
PSNR check, with GPU/Orin NX in FP32 and LENS in integer LSE.}
\label{tab:render_quality}
\centering
\setlength{\tabcolsep}{3pt}
\tighttablespacing
\begin{tabular*}{\columnwidth}{@{\extracolsep{\fill}}lcccc@{}}
\toprule
\multicolumn{5}{l}{\textit{Static-view crossing protocol: three-view mean}}\\
\midrule
Scene & ACEsplat & LENS & Lat.\,(ms) & perf/W \\
\midrule
ShopFacade    & 31.80 & 34.03 & 32.9 & 1.81x \\
KingsCollege  & 31.88 & 32.97 & 33.1 & 1.79x \\
OldHospital   & 31.82 & 32.33 & 32.0 & 1.86x \\
StMarysChurch & 33.92 & 34.46 & 33.7 & 1.76x \\
GreatCourt    & 36.57 & 36.75 & 27.7 & 2.14x \\
\midrule
mean          & 33.20 & \textbf{34.11} & 31.9 & \textbf{1.86x} \\
\midrule
\multicolumn{5}{l}{\textit{Reference-view platform check}}\\
\midrule
Scene & ACEsplat & GPU & Orin NX & LENS \\
\midrule
ShopFacade    & 31.80 & 33.55 & 33.55 & 32.09 \\
KingsCollege  & 31.88 & 33.61 & 33.61 & 33.09 \\
OldHospital   & 31.82 & 32.85 & 32.85 & 32.09 \\
StMarysChurch & 33.92 & 35.11 & 35.11 & 35.07 \\
GreatCourt    & 36.57 & 37.12 & 37.12 & 36.60 \\
\bottomrule
\end{tabular*}
\vspace{-3pt}
\end{table}

\begin{figure}[tbp]
\centering
\includegraphics[width=\columnwidth]{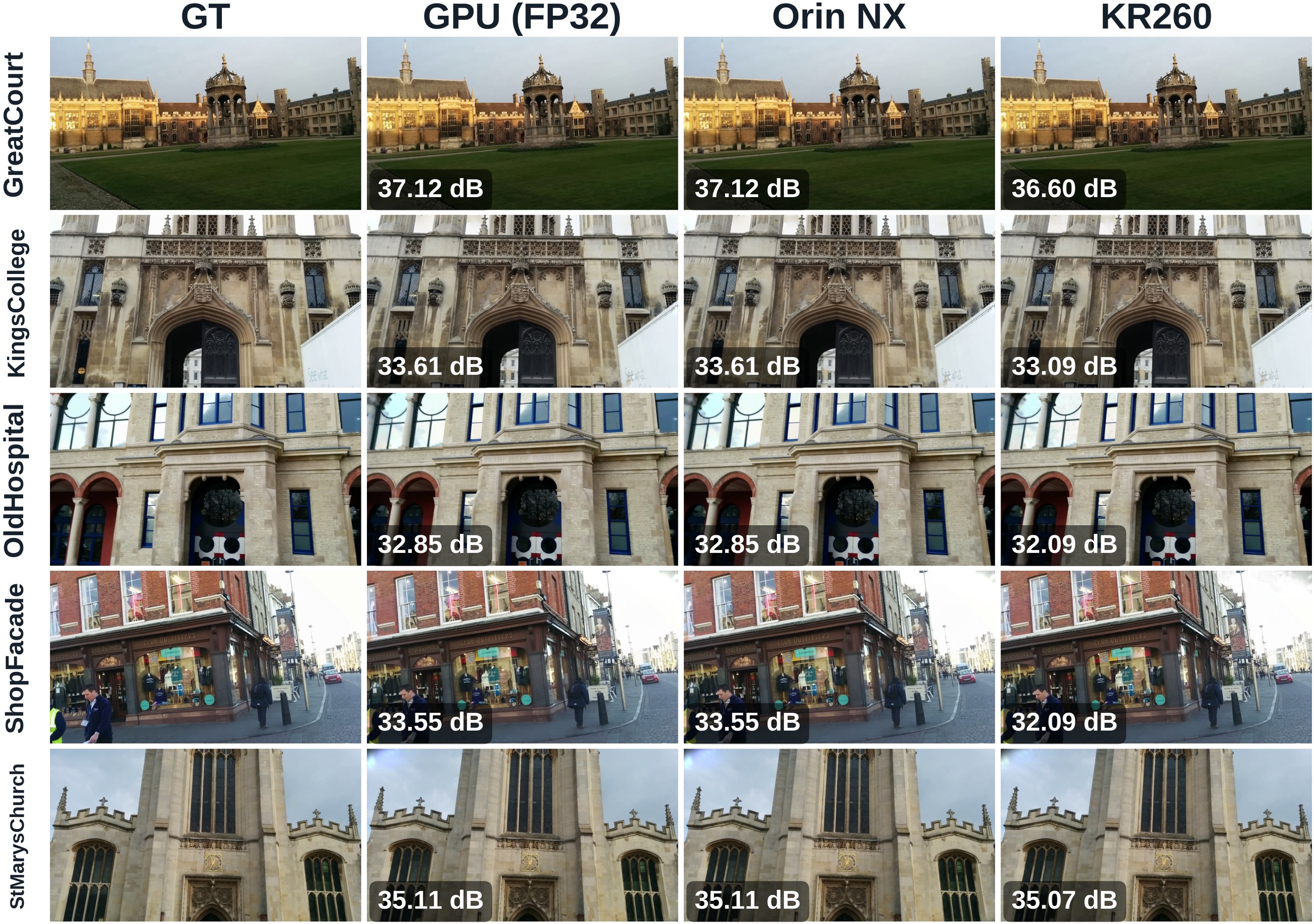}\\[2pt]
\includegraphics[width=\columnwidth]{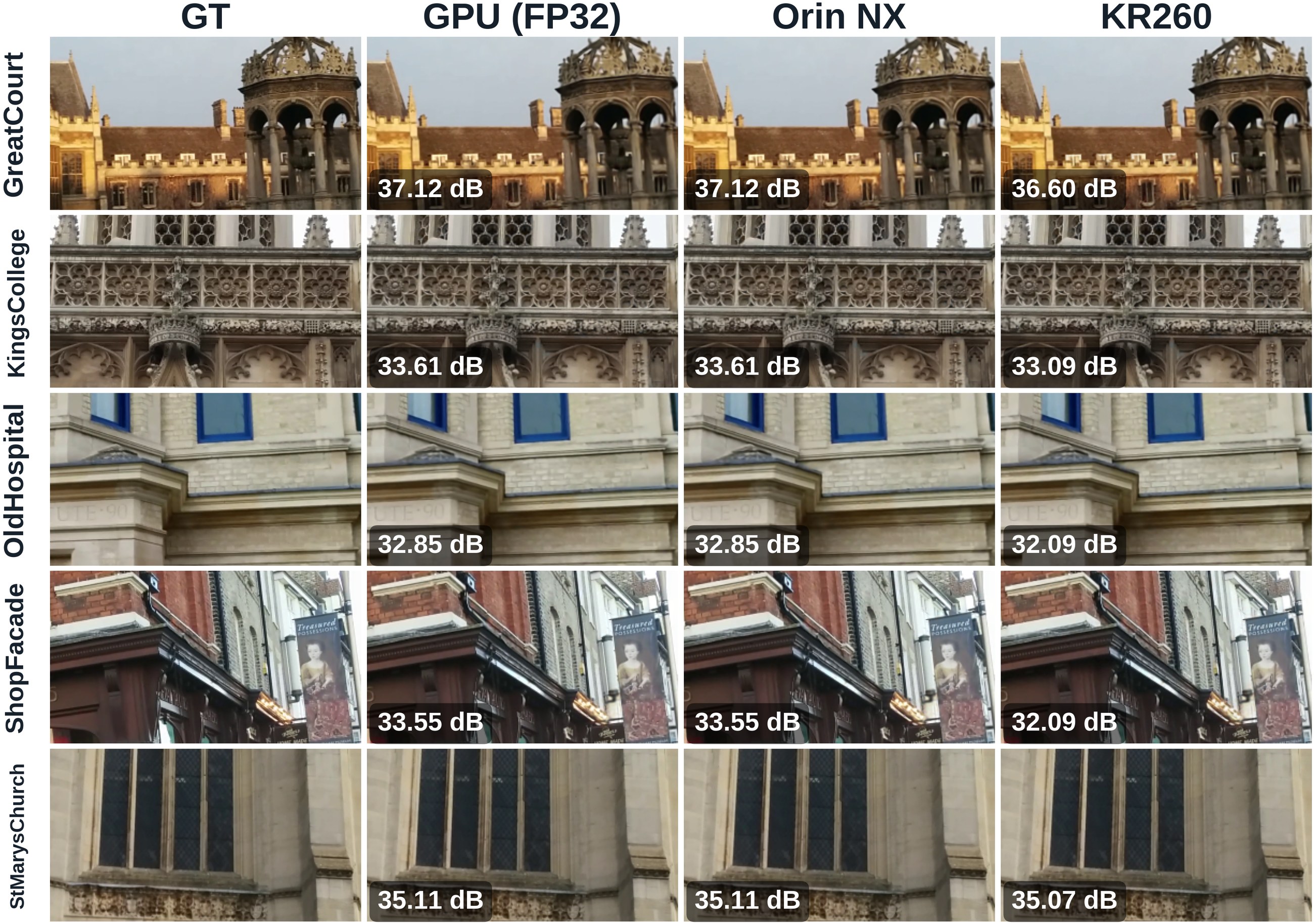}
\caption{Qualitative comparison on Cambridge Landmarks. \textit{Top:} full renders
(columns: ground truth, FP32 GPU, Jetson Orin NX, LENS integer render); the Orin NX
render is produced on the Jetson itself and the LENS column is bit-exact to the
integer emulator, with PSNR overlaid. \textit{Bottom:} detail crops of the same
region across the four columns, where the LENS render holds the fine structure of
the FP32 and Orin NX renders.}
\label{fig:qual_cambridge}
\vspace{-5pt}
\end{figure}

Every scene reaches the ACEsplat PSNR on board, for a three-view mean of 34.11 dB
against 33.20. The crossing protocol optimizes each field only to the first
iteration at which the on-board integer render reaches the scene's ACEsplat PSNR, so
the number is not inflated by over-optimization. Fig.~\ref{fig:qual_cambridge}
compares the ground truth, the FP32 GPU render, the Orin NX render, and the board
render; the board column is bit-exact to the integer emulator. The bottom panel
of Table~\ref{tab:render_quality} shows the platform check: GPU and Orin NX agree
to 0.01 dB with the float renderer, and the LENS integer renderer meets or
exceeds ACEsplat on every scene.

\subsection{Wayspots Rendering}

\begin{figure}[tbp]
\centering
\includegraphics[width=\columnwidth]{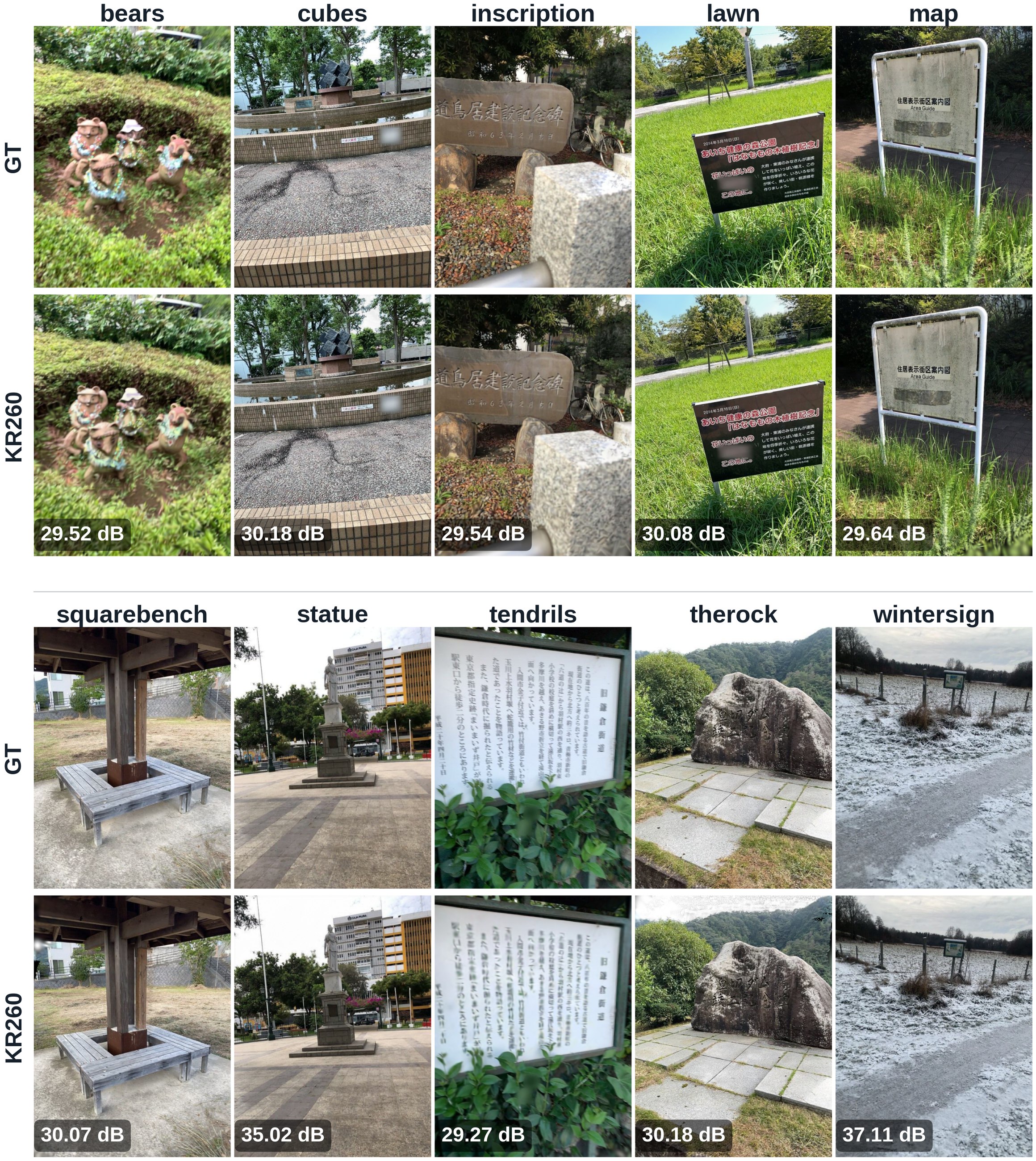}
\caption{Qualitative reconstruction rendering on all ten Wayspots scenes. Each
scene shows the ground truth and the LENS integer render (bit-exact to the
emulator), with the board PSNR overlaid. Every scene exceeds the 29.11 dB
ACEsplat average.}
\label{fig:qual_wayspots}
\vspace{-3pt}
\end{figure}

We render all ten Wayspots scenes on board (Fig.~\ref{fig:qual_wayspots}). Six
cross the ACEsplat average with SCR-point initialization; four require a
camera-ray grid initialization because SCR points pile into too few tiles in
high-frequency scenes (cubes, lawn) or degenerate-depth scenes (statue,
wintersign). Placing one Gaussian on the camera ray through each image-grid cell
restores coverage with a lean field. This lifts cubes, lawn, statue, and
wintersign above target without changing the render core, and the ten-scene
board mean is 31.06 dB against ACEsplat's 29.11 (Table~\ref{tab:wayspots_render}).

\begin{table}[tbp]
\caption{Wayspots reconstruction rendering PSNR (dB), board-measured and
bit-exact to the emulator. Grid-init scenes are marked $\dagger$.}
\label{tab:wayspots_render}
\centering
\setlength{\tabcolsep}{3pt}
\tighttablespacing
\begin{tabular*}{\columnwidth}{@{\extracolsep{\fill}}lclclc@{}}
\toprule
Scene & LENS & Scene & LENS & Scene & LENS \\
\midrule
bears        & 29.52 & squarebench    & 30.07 & cubes$^\dagger$      & 30.18 \\
inscription  & 29.54 & tendrils       & 29.27 & lawn$^\dagger$       & 30.08 \\
map          & 29.64 & therock        & 30.18 & statue$^\dagger$     & 35.02 \\
wintersign$^\dagger$ & 37.11 & \multicolumn{4}{c}{mean \textbf{31.06} dB} \\
\bottomrule
\end{tabular*}
\vspace{-5pt}
\end{table}

\subsection{Novel-View Synthesis}

\begin{figure}[tbp]
\centering
\includegraphics[width=\columnwidth]{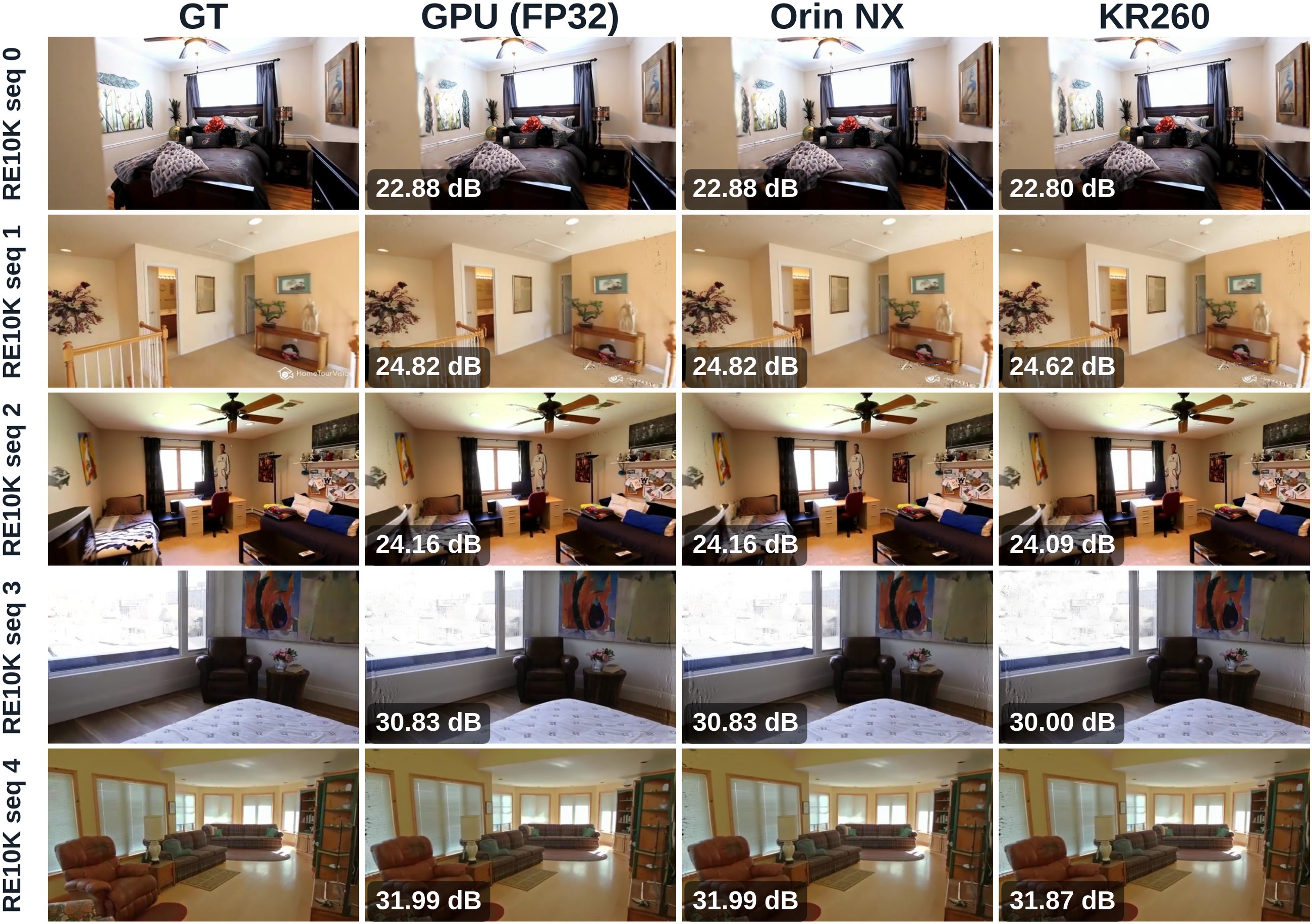}
\caption{Two-view novel-view synthesis on RealEstate10K. Columns: ground truth,
FP32 GPU render, Jetson Orin NX render, LENS integer render.}
\label{fig:qual_nvs}
\postfloatvspace
\end{figure}

For RealEstate10K two-view NVS, two input views are optimized off board and the
held-out target is rendered on the KR260. On fifteen targets the board mean is
27.84 dB against ACEsplat's 26.11, bit-exact to the emulator
(Table~\ref{tab:nvs}, Fig.~\ref{fig:qual_nvs}).

\begin{table}[tbp]
\caption{RealEstate10K two-view NVS PSNR (dB), board-measured and bit-exact to
the emulator.}
\label{tab:nvs}
\centering
\setlength{\tabcolsep}{3pt}
\tighttablespacing
\begin{tabular*}{\columnwidth}{@{\extracolsep{\fill}}lclclc@{}}
\toprule
Scene & LENS & Scene & LENS & Scene & LENS \\
\midrule
s0 & 22.80 & s5 & 30.13 & s10 & 26.53 \\
s1 & 24.62 & s6 & 28.46 & s11 & 32.49 \\
s2 & 24.09 & s7 & 31.63 & s12 & 26.08 \\
s3 & 30.00 & s8 & 26.23 & s13 & 20.86 \\
s4 & 31.87 & s9 & 32.16 & s14 & 29.67 \\
\midrule
\multicolumn{6}{c}{mean \textbf{27.84} dB} \\
\bottomrule
\end{tabular*}
\vspace{-5pt}
\end{table}

\subsection{On-Device Training Boundary}

Training is a boundary experiment. The Orin NX reference recipe converges in
41.5 s at 12.6 W; a four-A53 KR260 loop reaches 34.96 dB but takes 11.3 hours
(Fig.~\ref{fig:analysis}c). This confirms that ACEsplat-level quality is
reachable on the device, but deployment rendering does not make an efficient
training datapath.

\subsection{System-Level Analysis}

We close with the system-level limits exposed by the measurements.

\begin{figure*}[tp]
\centering
\includegraphics[width=\linewidth]{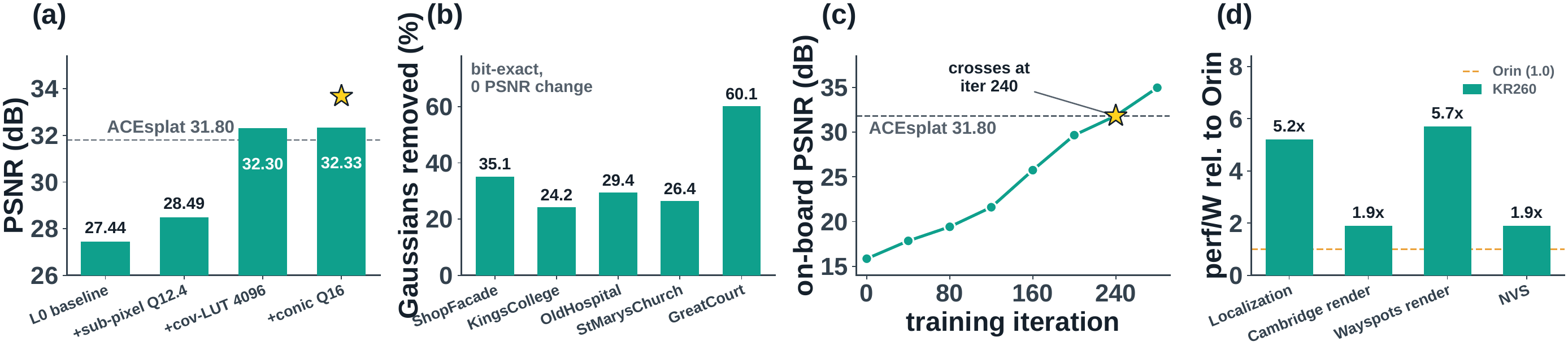}
\caption{Analysis. (a) Integer precision recipe on ShopFacade; the 4096-entry
covariance LUT is the dominant lever and the final configuration (star) clears
the ACEsplat PSNR. (b) Stream compaction removes 24 to 60 percent of streamed
Gaussians per scene with no change to any output bit. (c) On-device training
crosses the ACEsplat target at iteration 240 (star) and reaches 34.96 dB,
board-measured. (d) Performance per watt relative to the Jetson Orin NX across
deployment tasks; bars above the baseline favor LENS.}
\label{fig:analysis}
\vspace{-5pt}
\end{figure*}

\begin{figure}[tbp]
\centering
\includegraphics[width=\columnwidth]{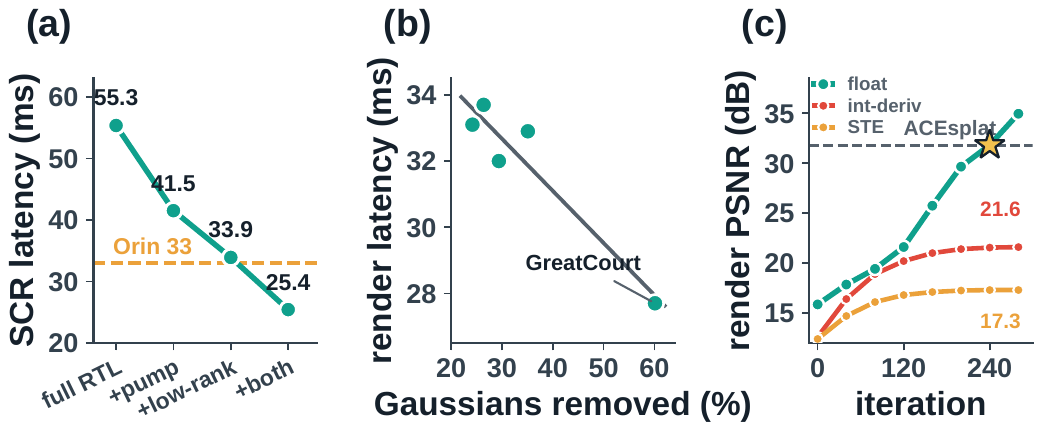}
\caption{Analysis line plots, board-measured. (a) SCR-forward latency across stacked
optimizations (full-rank RTL, then double-pumping, low rank, and both) falls from
55.3 to 25.4 ms, below the Jetson Orin NX (33 ms, dashed). (b) Render latency versus
the fraction of streamed Gaussians the compaction removes, per Cambridge scene;
GreatCourt sheds 60 percent and is the fastest. (c) On-device training boundary: a
float forward and backward converges and crosses the ACEsplat PSNR at iteration 240,
while differentiating the quantized render plateaus at 21.6 dB and a straight-through
estimator plateaus at 17.3 dB.}
\label{fig:analysis_lines}
\vspace{-5pt}
\end{figure}

\noindent\textbf{Performance per watt.}
LENS wins performance per watt over the Orin NX on every deployment task
(1.9x--5.7x, Table~\ref{tab:master}). The lean SCR and Wayspots workloads also win
absolute latency, while denser Cambridge and NVS renders are slower than Orin but
use much less power. The divider is Gaussian count per tile: latency falls with
stream compaction (Fig.~\ref{fig:analysis_lines}b), and SCR crosses under Orin
only after RTL systolic compute, double-pumping, and accuracy-safe low rank stack
together (Fig.~\ref{fig:analysis_lines}a). Thus the FPGA advantage is primarily
energy efficiency; it becomes a latency advantage when the Gaussian stream is
lean enough for the fabric datapath to stay fed without excessive memory traffic.

\noindent\textbf{On-device training boundary.}
Only a float forward/backward loop converges to 34.96 dB; differentiating the
deployment integer render plateaus at 21.6 dB, and a straight-through estimator~\cite{bengio2013ste}
plateaus at 17.3 dB (Fig.~\ref{fig:analysis_lines}c). The gradient implementation
matches autograd, so the limit is precision and silicon: a training accelerator
needs a smooth wide-precision datapath rather than the deployment renderer. This
is why training is reported as a boundary capability, not a deployment win.

\noindent\textbf{Capacity, precision, and initialization.}
Three effects can look like fixed-point error. The 4096-entry covariance LUT is
the main arithmetic lever (Fig.~\ref{fig:analysis}a); dense Cambridge scenes also
need the 4608-Gaussian tile cap; and the four hard Wayspots scenes need grid
initialization to avoid piling Gaussians into too few tiles. A lean, well-covered
field renders correctly in fixed point, while a dense or degenerate one does not.
The practical rule is to rule out capacity and initialization before blaming
integer arithmetic. This also explains why board measurement is necessary:
software replay without the hardware tile cap can make dense scenes look better
than the deployed renderer, while a bit-exact integer replay is faithful once the
same capped stream is used.

\noindent\textbf{Deterministic latency and energy.}
Over 25 ShopFacade runs the render averages 32.02 ms with a 32.25 ms maximum
(0.1\% CV and 0.7\% worst-case margin), a narrow spike rather than a tail of late
frames. The board draws 3.9 W, or about 0.125 J per frame versus roughly 0.24 J
on the Orin NX.

\section{Discussion and Limitations}

Taken together, the results show that LENS is best understood as a scene-memory
system, not a single generic accelerator. SCR benefits from systolic DSP
utilization, double-pumping, and accuracy-gated low-rank compression because its
acceptance criterion is pose accuracy. Rendering benefits from representation
and recurrence co-design, and from removing streamed work that provably cannot
change the output, because its limit is ordered Gaussian bandwidth and tile
capacity. This division is important for edge autonomy: the FPGA is not the best
device for every phase, but it is effective for the repeated localization and
rendering queries that dominate deployment. The reconstruction results use three views per scene, and NVS uses
fifteen held-out targets rather than the full RealEstate10K protocol; larger
protocol sweeps would tighten the averages. Cambridge rendering and NVS have
higher absolute latency than the Orin NX and win on performance per watt, not
wall-clock. On-device training reaches ACEsplat-level quality but not the GPU's
time or energy. Thus, LENS closes the deployment path for localization and
rendering, while leaving wide-precision training, a fused SCR-render-training
bitstream, and closed-loop on-robot deployment as future work.

\section{Conclusion}

LENS demonstrates that an embedded FPGA can serve an SfM-free scene memory when
localization, Gaussian rendering, and board data movement are co-designed rather
than accelerated in isolation. The SCR accelerator preserves ACEsplat-level pose
accuracy while reaching 25.42 ms on a KR260, and the LSE renderer runs bit-exact fixed-point
front-to-back 3D Gaussian rendering on the same platform, reaching
ACEsplat-level reconstruction, novel-view synthesis, and Wayspots rendering on
board at a performance-per-watt advantage over a Jetson Orin NX. On-device
training reaches ACEsplat-level quality but not the GPU's time or energy, which
marks the boundary where a dedicated wide-precision accelerator is still needed.
The result is an on-board deployment path for localization and rendering, and
a clear map of the remaining limits: precision, capacity, and data movement.

\bibliographystyle{IEEEtran}
\bibliography{refs}

\end{document}